\documentclass{bmvc2k}
\usepackage{graphicx,subfigure,multirow}

\title{Improved Techniques for the\\ Weakly-Supervised Object Localization}

\addauthor{Junsuk Choe}{skykite@yonsei.ac.kr}{1}
\addauthor{Joo Hyun Park}{pwangjoo@yonsei.ac.kr}{1}
\addauthor{Hyunjung Shim}{kateshim@yonsei.ac.kr}{1}

\addinstitution{
School of Integrated Technology\\
Yonsei University, South Korea
}

\runninghead{Choe, Park, Shim}{Improved WSOL: Under review at BMVC 2018}

\def\etal{\emph{et al}\bmvaOneDot}

\begin{document}

\maketitle

\begin{abstract}
We propose an improved technique for weakly-supervised object localization. Conventional methods have a limitation that they focus only on most discriminative parts of the target objects. The recent study addressed this issue and resolved this limitation by augmenting the training data for less discriminative parts. To this end, we employ an effective data augmentation for improving the accuracy of the object localization. In addition, we introduce improved learning techniques by optimizing Convolutional Neural Networks (CNN) based on the state-of-the-art model. Based on extensive experiments, we evaluate the effectiveness of the proposed approach both qualitatively and quantitatively. Especially, we observe that our method improves the Top-1 localization accuracy by 21.4 - 37.3\% depending on configurations, compared to the current state-of-the-art technique of the weakly-supervised object localization.
\end{abstract}

\section{Introduction}
\label{sec:intro}
Object localization aims to identify the location of an object in an image. The state-of-the-art object detection utilizes fully-supervised learning, which involves annotating locations, such as bounding boxes \cite{liu2016ssd,  redmon2016yolo, redmon2016yolo9000, ren2017faster}. Unfortunately, rich annotations involves intensive manual tasks, and are often quite different from different human participants. Weakly supervised approaches to the object localization bypass the issue of annotating localization, with only image-level labels. Because weakly supervised approaches do not rely on the bounding box annotation, they can be a practical alternative.

Existing approaches can be categorized based on whether they derive discriminative features of the training dataset explicitly or implicitly. Explicit methods utilize handcrafted features to extract class-specific patterns for object localization \cite{weber2000unsupervised, fergus2003object, bilen2014weakly, song2014learning, cinbis2014multi, song2014weakly, cinbis2017weakly}. Meanwhile, implicit methods first train deep convolutional neural networks (CNN) mostly for object classifications using image-level labels, then utilize its byproduct, the activation map, for the object localization. The final heatmap can be produced by leveraging the activation maps from the networks. Applying the simple post-processing on the extracted heatmap, it is possible to localize a target object \cite{simonyan2013deep, oquab2015object, zhou2016learning}. Both approaches, however, result in capturing most discriminative parts for object localization, discarding less discriminative parts. This causes the bounding box to lack coverage among entire parts of the object.

\begin{figure}[t]
\centering
\includegraphics[width=.8\columnwidth]{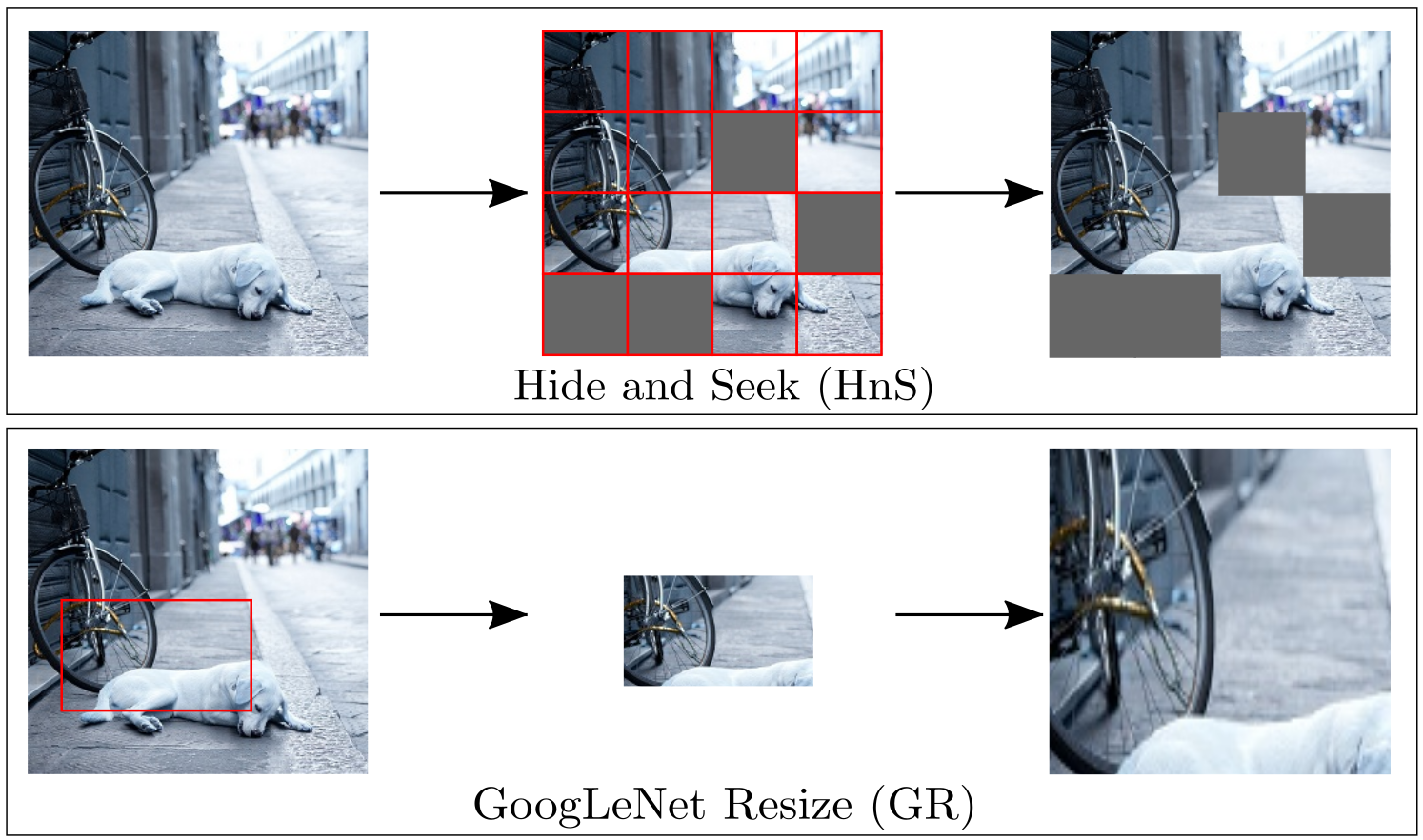}
\caption{Examples of augmentation methods of Hide-and-Seek (HnS) and GoogLeNet Resize (GR). HnS divides an image into grids, randomly removing some patches. GR randomly crops the input image into a rectangular patch, then resize it to the original input size.}
\label{fig:augments}
\end{figure}

Recently, to resolve such shortcomings, several techniques have been proposed \cite{singh2017hide, kim2017two, zhang2018adversarial, li2018tell}. Singh and Lee \cite{singh2017hide} suggested a new training technique, namely \emph{Hide-and-Seek} (HnS), that involves a grid mask for introducing obscured training data. More specifically, they randomly hide the sub-regions of input image by eliminating each patch of the grid mask with a prefixed probability. Their method potentiates the possibility to \emph{hide} the most discriminative parts of the object, allowing the CNN to \emph{seek} the less discriminative part. Their approach can be interpreted as the data augmentation technique in a sense that only the training dataset is being modified \cite{li2018tell}, leading to an advantage that holds independence to a specific classification algorithm. Based on quantitative and qualitative evaluations, \cite{singh2017hide} achieved the state-of-the-art performance among weakly-supervised object localization techniques.

In this paper, we suggest a data augmentation and improved training techniques that is effective on increasing the accuracy of the weakly-supervised object localization. We construct a baseline network with the state-of-the-art classification network \cite{he2016identity}. To produce a heatmap, the Class Activation Maps (CAM) algorithm \cite{zhou2016learning} is applied. Then, the baseline network outputs classification (i.e., object labels) and localization (i.e., bounding box) results. We investigate two aspects for improving the localization performance: 

\begin{enumerate}
  \item How to further improve the data augmentation proposed by Hide-and-Seek \cite{singh2017hide}?
  \item How network capacity and batch size influence the localization performance?
\end{enumerate}

In investigation of the data augmentation, we have found that a GoogLeNet Resize (GR) augmentation method \cite{szegedy2015going} can improve the localization performance even better than the state-of-the-art technique (HnS). As shown in Figure \ref{fig:augments}, both GR and HnS deal with how to increase the training dataset for encapsulating less discriminative parts of object. GR uses the partial regions of image, while HnS hides partial regions. Although both methods aim to include less discriminative part of the object in the bounding box, GR is inherently more aggressive in that it augments more challenging images with small, and valid regions. From the experimental evaluation, we show that GR outperforms HnS in weakly-supervised object localization. 

For the improved training schemes, we investigate the effect of batch sizes and depths of a network. First, inspired by recent report \cite{li2017visualizing} that the size of the minibatch has critical influence on CNN optimization, we examine the influence of the batch size on the performance of weakly-supervised object localization. With empirical studies, we find that smaller batch size allows better performance. Secondarily, the network capacity can also influence the localization performance. We compare the network depths with 18-layer and 34-layer. With such comparison, we observe that deeper network performs better in the object localization. 

For evaluation, we use Tiny ImageNet dataset, a reduced version of ImageNet \cite{russakovsky2015imagenet}. Notice that the Tiny version is more challenging for conducting a localization task than the original ImageNet due to the reduced resolution and the lack of training data. Finally, by integrating our improved training techniques, we have improved the localization performance (Top-1 localization accuracy \cite{singh2017hide}) from 27.1\% to 36.0\%, which is approximately 30\% improvement over the current state-of-the-art technique \cite{singh2017hide}. 

\section{Algorithm Description}
\label{sec:approaches}

In this section, we first explain the baseline algorithms, and then show the effect of the data augmentation, minibatch size, and network depth on the performance of object localization. 

\textbf{Baseline algorithm.} We implement the classification network using the pre-activation residual network \cite{he2016identity}. To produce the heatmaps, we use the Class Activation Maps (CAM) method \cite{zhou2016learning} for extracting the heatmap from the classification network. Note that the pre-activation residual network that we choose is a state-of-the-art classifier, which is an improved version of ResNet \cite{he2016deep}. By conducting the large-scale experiments, we have found that the pre-activation ResNet based CAM is superior to original CAM, which is built based on GoogLeNet \cite{szegedy2015going} or VGG \cite{simonyan2014very} architecture. The CAM replaces the fully connected layer right after the last convolutional layer with the Global Average Pooling (GAP) layer, which the scheme is applicable to any type of CNN networks. With the GAP layer, the spatial information of the feature map is visualized. We then obtain the heatmap by aggregating the higher-layer activation maps with the weights between GAP layer and softmax layer. Following \cite{zhou2016learning}, the final output, a bounding box, is obtained by thresholding the heatmap. 

\textbf{Motivation.} Hide-and-Seek (HnS) \cite{singh2017hide} is one of the state-of-the-art techniques in weakly-supervised object localization. Note that previous methods only focuses on the most discriminative parts. Consequently, localization results from previous work tend to produce a smaller bounding box, ignoring less discriminative parts. To resolve this problem, HnS aims to learn less discriminative parts of the object by hiding random parts of the object. Specifically, HnS first divides an input image into small patches in a grid-style. HnS then randomly selects several patches in the grid to mask the patch. With such random masking, it is more likely to hide the most discriminative parts of the object; the network is more likely to learn less discriminative parts of the object. By doing this, \cite{singh2017hide} declares that HnS can achieve the state-of-the-art performance. 

The key idea of HnS is to hide the most discriminative parts of target object from CNN so that the network is also capable of learning the less discriminative parts of object. Motivated by this idea, we adopt the GoogLeNet Resize (GR) augmentation \cite{szegedy2015going}. GR augmentation randomly crops 8-100\% of an input image with the aspect ratio between 0.75 and 1.33, and then resizes the cropped image to the original input image size. Figure \ref{fig:augments} visualizes the HnS and GR augmentation respectively. GR augmentation effectively produces the training data at various scales (i.e., various levels of zoom-in). In this way, we can exam the local structure as well as the global structure during training. As a result, patches drawn from GR augmentation enforce the classifier to learn the less discriminative local structure better than HnS, which only observes the global structure (i.e., a single scale) during training. GR augmentation is analogous with HnS in that it only provides partial information of an object to a CNN. However, we expect that the GR augmentation, which only utilizes partial parts of an object, is more robust to deformations and pose variations than HnS, which only hides partial parts of an object. Based on empirical study, we exam which one is superior, and whether two methods are complementary or mutually exclusive.

\textbf{Batch size.} Recently, Li \etal \cite{li2017visualizing} claimed that the smaller the batch size we use, the closer the CNN loss landscape approaches to the convex function; reaching the global optima more easily. However, the CNN classifier with the batch normalization (BN) \cite{ioffe2015batch} requires the large batch size. Note that BN computes the local mean and variance within the minibatch. Hence, it is more likely to have the mismatch between the local and global statistics (i.e., mean and variance) when the batch size is small. Likewise, it is still controversial how batch sizes affect the actual performance of a CNN \cite{chaudhari2016entropy, keskar2016large, dinh2017sharp}. Heretofore, there is no consensus of what batch size is optimal for object localization. To reveal the effect of batch size in our application, we examine the performance of the object localization by varying the batch sizes. 

\textbf{Network depth.} Generally, the classification performance of deep neural networks outperforms that of a shallow neural network \cite{krizhevsky2012imagenet, simonyan2014very}. Meanwhile, if depth is too deep, the classification performance decreases due to gradient vanishing. This problem is well addressed by \cite{he2016deep, li2017visualizing}. We overcome such problem by utilizing an identity mapping \cite{he2016identity}, thereby making it possible to increase the depth of network without gradient vanishing. Note that all these discussions regarding the effect of network depth were made on the classification tasks. Hence, we empirically analyze how the performance of the object localization is influenced with the network depths.

\section{Experimental Results}
\label{sec:experiments}

In this section, we first describe the implementation detail, and then show how the data augmentation, batch size, and the network depth influence the accuracy of weakly-supervised object localization. In addition, we also show qualitative evaluation results as well. 

\textbf{Implementation details.} We utilized Tiny ImageNet dataset for training the classification network. The Tiny ImageNet is simplified version of the ImageNet \cite{russakovsky2015imagenet}, with the image size reduced to $64\times64$. There are total 200 categories, 500 training images, and 50 validation images per each category in the original Tiny ImageNet. For object recognition or localization tasks, handling the Tiny ImageNet is more challenging than the original ImageNet in two perspectives: the resolution and the volume of dataset. The resolution of the Tiny ImageNet is lower than that of ImageNet images, approximately one sixth. Previous study \cite{odena2016conditional} pointed out that low-resolution image is more difficult to recognize than high-resolution image. In addition, the number of data in ImageNet is two order of magnitude more than that of Tiny ImageNet. It is widely known that the larger dataset improves recognition performance. 

We use their 50 validation images as our test dataset. As we described in Section \ref{sec:approaches}, we use pre-activation ResNet for classification network, with slight modification to the size of input layer. We train the network using Nesterov momentum optimizer \cite{bengio2013advances} for 1500 epochs, and set the momentum to 0.9. The initial learning rate is 0.1 and we reduce the learning rate by a factor of 10 every 250 epochs. The weight decay is 1e-4. We utilize Tensorflow \cite{abadi2016tensorflow}, and Tensorpack \cite{wu2016tensorpack} for implementation of the codes. The HnS grid is implemented by following the mixed method, described further in their paper. The grid size [$0\times0$, $4\times4$, $8\times8$, $16\times16$] is randomly applied to the input image at every iteration.

\textbf{Evaluation metrics.} For evaluation, we utilize the same metrics used in \cite{singh2017hide}:
\begin{enumerate}
  \item Top-1 localization accuracy (\textit{Top-1 Loc}): The ratio of samples, which intersection over union (IoU) between estimates and ground truth is more than 50\% and, at the same time, classification result is correct.
  \item Localization accuracy with known ground truth class (\textit{GT-known Loc}): The ratio of samples, which IoU between estimates and ground truth is more than 50\%.
  \item Top-1 classification accuracy (\textit{Top-1 Clas}): The ratio of correct answers among all test samples. 
\end{enumerate}

\begin{figure}[!t]
    \begin{center}
        \subfigure[CAM \cite{zhou2016learning}]{\includegraphics[width=0.28\columnwidth]{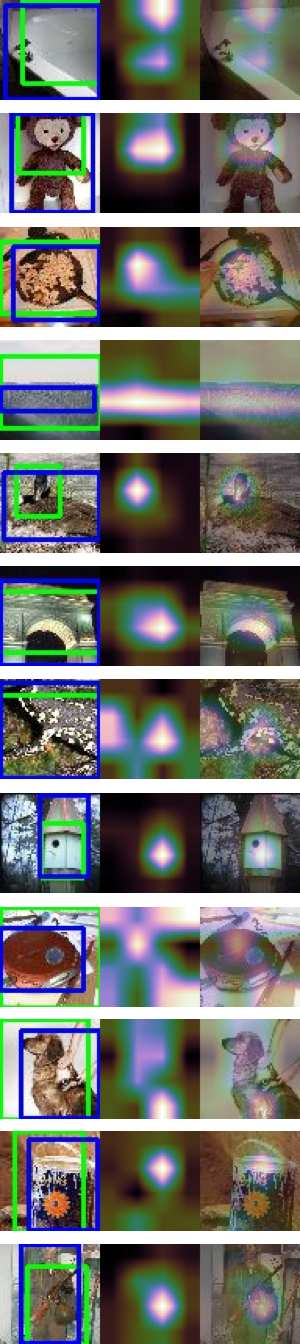}}
        \subfigure[HnS \cite{singh2017hide}]{\includegraphics[width=0.28\columnwidth]{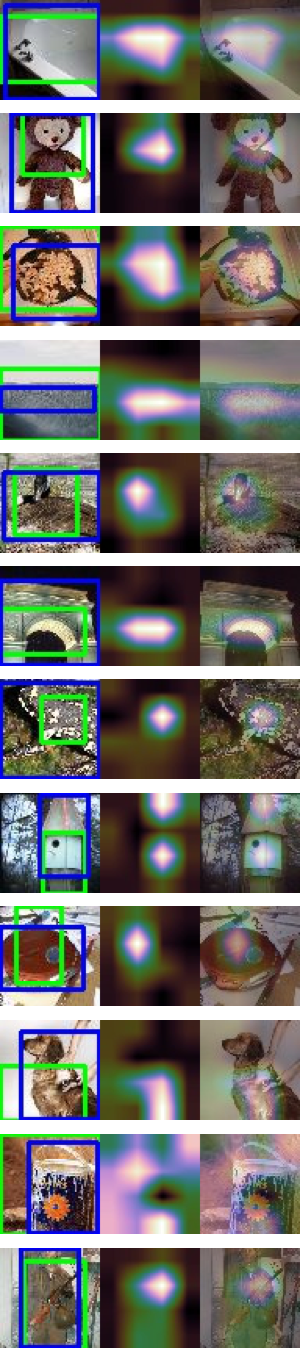}}
        \subfigure[GR (Proposed)]{\includegraphics[width=0.28\columnwidth]{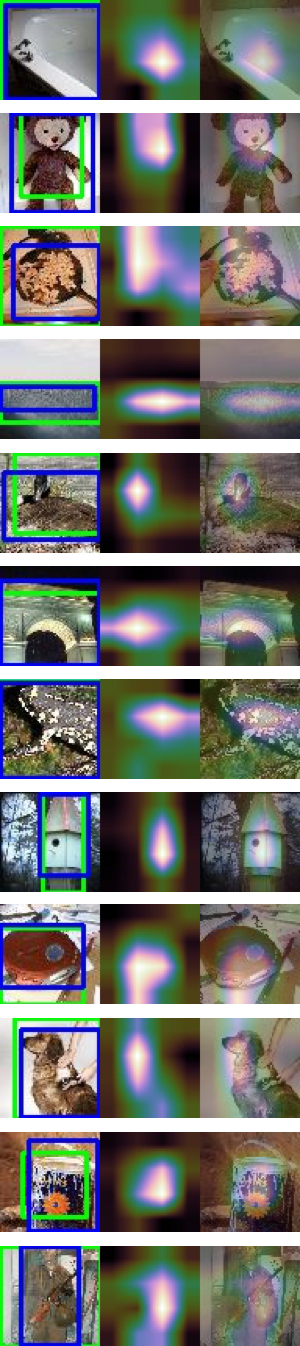}}
    \end{center}
    \caption{Qualitative evaluation results. GR experimental results clearly show better localization performance than the baseline and HnS. The blue bounding box is a ground truth while the green bounding box is an estimate. Left: input image. Middle: heatmap. Right: input image overlapping with heatmap.}
\label{fig:qualitative}
\end{figure}

\textbf{Quantitative evaluation.} We compare the performance of object localization for HnS and the proposed algorithm utilizing GR augmentation. In addition, we sequentially apply both HnS and GR augmentation and observe the effect of the sequential order of HnS and GR to the performance. We set the batch size to 256 in all data augmentation experiments for a fair comparison. Table \ref{tab:augmentation} shows the experimental results. We highlight superior results with bold letters. From these results, we observe that both HnS and GR performances are better than the baseline (CAM without data augmentation). Among HnS and GR, we conclude that GR clearly outperforms HnS in all three metrics. More specifically, the \textit{GT-known Loc} of our results outperforms the HnS by 7\% and the CAM by 8\%. In addition, our results outperform the HnS by almost 25\% with a \textit{Top-1 Loc} metric. The \textit{Top-1 Clas} performance of our algorithm is better than that of HnS and CAM. Lastly, we find that the performance decreases when HnS and GR are simultaneously applied, meaning that HnS and GR are mutually exclusive. From these observations, we finally conclude that solely applying GR is the better choice to improve performance.

\begin{table*}[!t]
\centering
\label{tab:augmentation}
\begin{footnotesize}
\begin{tabular}{cc|ccccc}

\hline
Depth                     & Metric       & CAM \cite{zhou2016learning} & HnS \cite{singh2017hide} & GR (Proposed)          & HnS after GR & GR after HnS \\ \hline
\multirow{3}{*}{ResNet34} & GT-known Loc & 53.46    & 54.30    & \textbf{57.82} & 57.49        & 57.43        \\
                          & Top-1 Loc    & 27.50    & 29.72    & \textbf{36.00} & 33.93        & 33.45        \\
                          & Top-1 Clas   & 44.54    & 46.67    & \textbf{54.13} & 51.28        & 50.45        \\ \hline
\multirow{3}{*}{ResNet18} & GT-known Loc & 53.42    & 52.89    & \textbf{57.00} & 56.99        & 56.76        \\
                          & Top-1 Loc    & 27.13    & 27.56    & \textbf{34.43} & 33.33        & 32.28        \\
                          & Top-1 Clas   & 43.90    & 44.12    & \textbf{52.23} & 50.14        & 48.87        \\ \hline

\end{tabular}
\end{footnotesize}
\caption{Accuracy comparison with HnS and various data augmentation techniques. The batch size is 256.}
\end{table*}

\textbf{Effect of batch size and network depth.} Next, we examine how the batch size and network depth affect object localization performance. Note that our hyper-parameters are tuned for batch-256 and 18-layer settings, and fixed for all other experiments. Table \ref{tab:learning} shows the experimental results. We again use bold letters for superior results. The performance is improved consistently by 10-15\% in all three methods. Although the hyper-parameters are not tuned for smaller batch size and deeper network, the experimental results clearly demonstrate that smaller batch size and deeper network produce higher accuracy in Top-1 localization (\textit{Top-1 Loc}). From this experimental study, we can conclude that smaller batch size and deeper network increase the accuracy of weakly-supervised localization.

\begin{table}[t]
\centering
\begin{footnotesize}
\begin{tabular}{cc|cc}
\hline
Method                    & Batch size & ResNet34          & ResNet18          \\ \hline
\multirow{3}{*}{CAM \cite{zhou2016learning}} & 32         & \textbf{31.49}    & 28.47             \\
                          & 128        & 29.62             & 27.76             \\
                          & 256        & 27.50             & 27.13             \\ \hline
\multirow{3}{*}{HnS \cite{singh2017hide}} & 32         & \textbf{31.17}    & 29.45             \\
                          & 128        & 30.32             & 29.25             \\
                          & 256        & 29.72             & 27.56             \\ \hline
\multirow{3}{*}{GR (Proposed)}  & 32         & \textbf{37.84}    & 35.94             \\
                          & 128        & 36.13             & 35.83             \\
                          & 256        & 36.00             & 34.43             \\ \hline
\end{tabular}
\end{footnotesize}
\caption{Top-1 localization accuracy of object localization upon various batch size and network depth.}
\label{tab:learning}
\end{table}

\textbf{Qualitative evaluation.} Lastly, we visually compare our best model (i.e., GR augmentation with batch-32 and 34-layer) in quantitative experiments with the baseline and HnS. Figure \ref{fig:qualitative} shows the experimental results. The left image shows estimated (green color) and ground truth (blue color) bounding box. The heatmap is shown in middle. The right image shows an input image overlapping with the heatmap. Note that the bounding box is obtained by post-processing the heatmap as proposed by CAM \cite{zhou2016learning}.

The qualitative evaluation results clearly show that our results can capture the entire parts of object better than CAM and HnS. As discussed by HnS \cite{singh2017hide}, CAM concentrates only on the most discriminative parts so to localize the partial region of target object. This issue is alleviated both HnS and the proposed approach; the combination of GR augmentation with deeper network and small batch size. Compared to HnS, our results can better cover the overall parts of target object, thus much closer to the ground truth bounding box. The same observation holds in heatmaps. For example, our results for heatmaps are more accurate than those of CAM and HnS, both in terms of the object coverage and position. Note that the green box is sometimes invisible in our results. It is because the estimated bounding box completely overlaps with the ground truth.

\section{Conclusions}
\label{sec:conclusions}
In this paper, we improve the performance of the weakly-supervised object localization in three aspects: the data augmentation, batch sizes, and the network depths. It is experimentally shown that the GoogLeNet Resize augmentation is better than the current state-of-the-art data augmentation technique \cite{singh2017hide} for weakly-supervised object localization. We also show that the performance increases with a small batch size. Finally, we show that deeper network produces better performance than shallower networks. We train pre-activation ResNet using Tiny ImageNet, and evaluate our methods both quantitatively and qualitatively. In the future, we aim to study new data augmentation methods to ensure better performance than GoogLeNet Resize augmentation, and analyze the performance saturation upon the network depths and the batch sizes for the weakly-supervised object localization.

\end{document}